\documentclass[letterpaper, 10 pt, conference]{ieeeconf}  

\IEEEoverridecommandlockouts                              

\usepackage{amsmath,amsfonts,amssymb}
\usepackage{algorithmic}
\usepackage{algorithm}
\usepackage{array}
\usepackage{bbm}
\usepackage[caption=false,font=footnotesize,labelfont=footnotesize,textfont=footnotesize]{subfig}
\usepackage{textcomp}
\usepackage{stfloats}
\usepackage{url}
\usepackage{verbatim}
\usepackage{graphicx}
\usepackage{booktabs}
\usepackage{color}
\usepackage[dvipsnames]{xcolor}
\definecolor{table_c}{RGB}{255,255,240}
\usepackage{diagbox}
\usepackage{multirow}
\usepackage{cite}
\usepackage{flushend}
\usepackage{threeparttable}
\makeatletter
\let\NAT@parse\undefined
\makeatother
\usepackage{hyperref}
\usepackage{balance}
\hypersetup{pdfstartview=FitH,
            colorlinks=true,
            linkcolor=red,
            anchorcolor=blue,
            citecolor=green
            }

\title{\LARGE \bf
DaDiff: Domain-aware Diffusion Model for Nighttime UAV Tracking
}

\author{Haobo Zuo$^{1}$, Changhong Fu$^{2,*}$, Guangze Zheng$^{1}$, Liangliang Yao$^{2}$, Kunhan Lu$^{2}$, and Jia Pan$^{1}$
\thanks{*Corresponding author}
\thanks{$^{1}$Haobo Zuo, Guangze Zheng, and Jia Pan are with the Department of Computer Science, University of Hong Kong, and also with the Centre for Transformative Garment Production, Hong Kong 999077, China.}%
\thanks{$^{2}$Changhong Fu, Liangliang Yao, and Kunhan Lu are with the School of Mechanical Engineering, Tongji University, Shanghai 201804, China. Email: changhongfu@tongji.edu.cn}%
}

\begin{document}

\maketitle
\thispagestyle{empty}
\pagestyle{empty}

\begin{abstract}

Domain adaptation is an inspiring solution to the misalignment issue of day/night image features for nighttime UAV tracking. However, the one-step adaptation paradigm is inadequate in addressing the prevalent difficulties posed by low-resolution (LR) objects when viewed from the UAVs at night, owing to the blurry edge contour and limited detail information. Moreover, these approaches struggle to perceive LR objects disturbed by nighttime noise. To address these challenges, this work proposes a novel progressive alignment paradigm, named domain-aware diffusion model (\textbf{DaDiff}), aligning nighttime LR object features to the daytime by virtue of progressive and stable generations. The proposed DaDiff includes an alignment encoder to enhance the detail information of nighttime LR objects, a tracking-oriented layer designed to achieve close collaboration with tracking tasks, and a successive distribution discriminator presented to distinguish different feature distributions at each diffusion timestep successively. Furthermore, an elaborate nighttime UAV tracking benchmark is constructed for LR objects, namely NUT-LR, consisting of 100 annotated sequences. Exhaustive experiments have demonstrated the robustness and feature alignment ability of the proposed DaDiff.
The source code and video demo are available at \url{https://github.com/vision4robotics/DaDiff}.

\end{abstract}

\section{Introduction}

Vision-based UAV tracking has been widely applied for intelligent robot applications, \textit{e.g.}, motion object analysis \cite{8736008}, 
geographical survey \cite{5942155}, 
and visual localization \cite{9157450}. 
With high-quality daytime tracking datasets \cite{8922619,lin2014microsoft,russakovsky2015imagenet}, the state-of-the-art (SOTA) trackers~\cite{9577739,9157457} have achieved superior performance. However, these trackers perform poorly in night scenes because of the decreased illumination, signal-to-noise ratio, and contrast of nighttime images compared to daytime ones \cite{9696362,9636680}. The above differences between day and night images or image features cause the distribution discrepancy, spawning an extremely challenging application, \textit{i.e.}, nighttime UAV tracking \cite{Li_2021_ICRA}.

In literature, the SOTA methods \cite{9696362,9636680} construct tracking-oriented low-light enhancers with cutting-edge trackers to realize nighttime UAV tracking. Nevertheless, this kind of plug-and-play method generally focuses on the image level and can scarcely learn to minimize the distribution gap at the feature level, which is insufficient in providing discriminative image features required for high-accuracy tracking. Although the one-step adaption paradigm~\cite{9879981} is researched to achieve image feature alignment with end-to-end training, 
\begin{figure}[t]
\centering
	{	
 \includegraphics[width=1.0\linewidth]{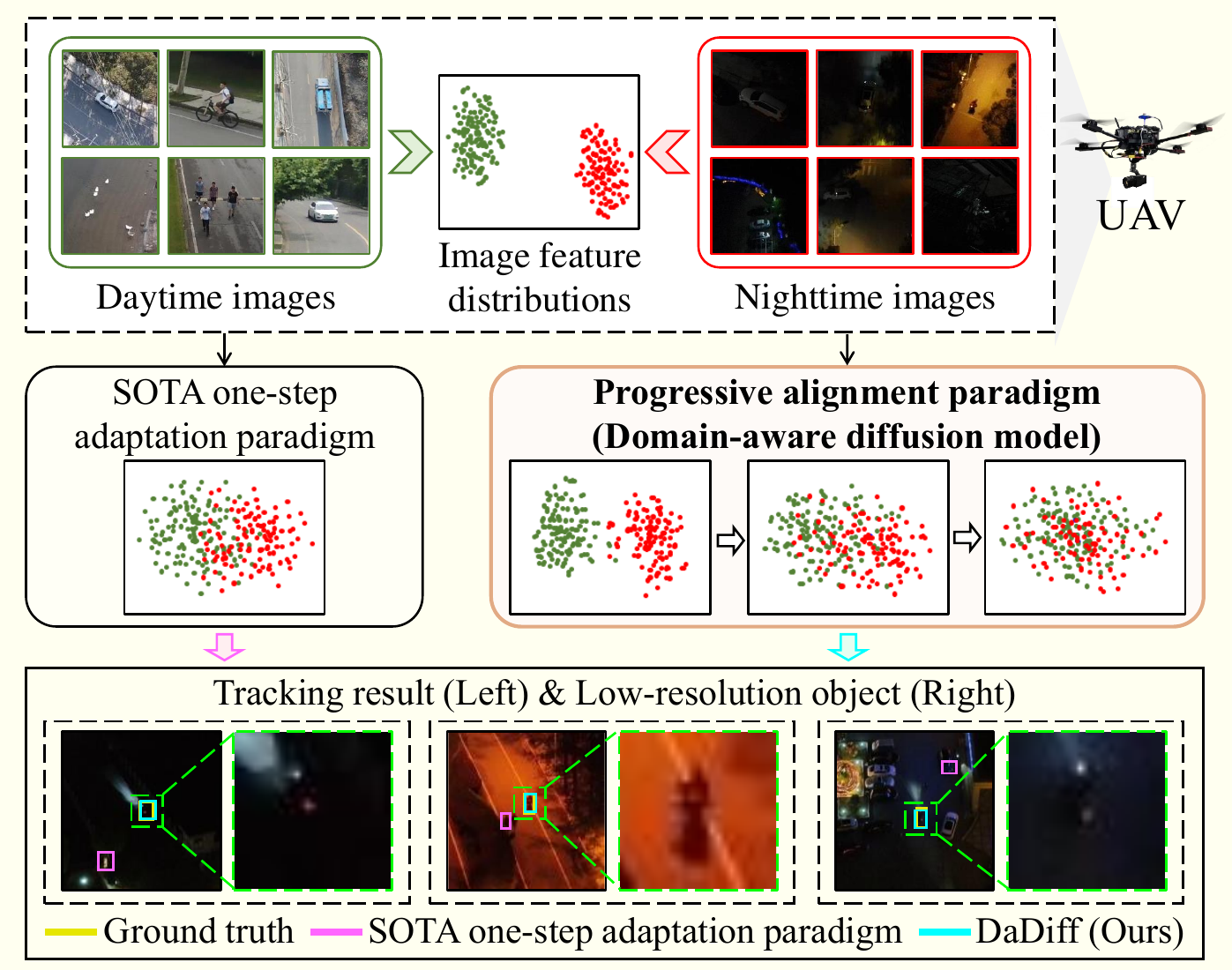}
	\centering
}
	\vspace{-15pt}
	\caption
	{
		Comparison of the one-step adaptation paradigm and the proposed domain-aware diffusion model, \textit{i.e.}, DaDiff, for nighttime UAV tracking. The feature distributions are visualized through t-SNE \cite{van2008visualizing}. \textcolor[rgb]{0.33,0.51,0.21}{Green} and {\color{red}red} indicate the daytime and nighttime image feature distributions, respectively. The scattergrams depict day/night feature distributions from different feature alignment methods. DaDiff successively and steadily narrows feature distribution discrepancy, thereby achieving superior tracking results, especially for low-resolution (LR) objects.
	}
	\label{1}
	\vspace{-15pt}
\end{figure}
such kind of method performs unstably when facing common low-resolution (LR) object challenges from UAV perspectives due to the following two reasons: \textit{1)} LR objects are hard to be identified from the background in one step due to the limited detail information and nighttime noise interference \cite{9879411}; \textit{2)} aligning the features of nighttime LR objects in one step is unstable due to the mismatch between the receptive field on LR features and the object sizes \cite{9010716}. \textbf{\textit{Therefore, how to align the nighttime LR object features to the daytime effectively and stably for robust nighttime UAV tracking is an urgent problem.}}

Diffusion models \cite{ho2020denoising,dhariwal2021diffusion,song2020denoising} have achieved superior performance in reconstructing the object information for LR images \cite{gao2023implicit,li2022srdiff,yue2024resshift,9887996}.
Diffusion-based methods can be regarded as a sort of variable model that uses a Markov chain to convert noise into data distribution. 
The generation ability of these models is typically derived from the step-by-step closing to the data distribution with the U-Net neural network. Additionally, such formulation allows for a guiding mechanism to control the image generation process with stability~\cite{yang2022diffusion}. 
Compared to one-step generation methods~\cite{9555209,9879981}, multi-step diffusion models \cite{song2020denoising} offer significant advantages. They can avoid issues such as high-frequency information loss, excessive smoothness, mode collapse, and effect instability~\cite{li2022srdiff}. By progressively enhancing detail information and sharpening edge contours of LR objects, diffusion models show promise for addressing feature alignment issues in the common LR object challenges of nighttime UAV tracking. Despite their potential, diffusion models have not yet been explored for nighttime UAV tracking. Furthermore, diffusion models are trained independently and cannot be seamlessly integrated into downstream tracking tasks. Therefore, bridging strategies are needed to leverage these models for nighttime UAV tracking effectively.

This work introduces the diffusion models into nighttime UAV tracking for the first time, proposing a novel domain-aware diffusion model, \textit{i.e.}, DaDiff. Specifically, the alignment encoder is developed to obtain valid domain-aware information of LR objects in negative light conditions. The tracking-oriented layer is presented to achieve close collaboration with tracking. To ensure the stability of alignment, the successive distribution discriminator is applied for identifying the different feature distributions at each diffusion timestep. The aligned result comparison of DaDiff and the one-step adaptation paradigm is exhibited in Fig. \ref{1}. DaDiff raises the tracking performance through successive and stable feature alignment. Besides, NUT-LR, an elaborate nighttime tracking benchmark, is constructed including 100 annotated sequences as the first LR object benchmark for nighttime UAV tracking. It focuses on the LR object challenges of nighttime UAV tracking, aiming at promoting the research on nighttime tracking to a broader field. The main contributions of this work are as follows:
\begin{itemize}
\item A novel progressive alignment paradigm, \textit{i.e.}, DaDiff, is proposed for nighttime UAV tracking. According to our knowledge, this work first applies diffusion models for nighttime UAV tracking.
\item An alignment encoder is developed to strengthen the detail information of nighttime LR objects. A tracking-oriented layer and a successive distribution discriminator are included to closely connect with the tracking tasks and gradually narrow the feature distribution gap between daytime and nighttime. 
\item A pioneering benchmark namely NUT-LR, comprising 100 annotated sequences with LR object challenges, is constructed for evaluation of LR object nighttime tracking under the UAV perspective. 
\item Comprehensive evaluation on NUT-LR, NUT-L \cite{yao2023sam}, and UAVDark70 \cite{Li_2021_ICRA} benchmarks demonstrate the effectiveness and feature alignment ability of the proposed DaDiff for nighttime UAV tracking.
\end{itemize}


\section{Related work}

\subsection{Nighttime UAV tracking} 

Nighttime UAV tracking has been applied for numerous practical applications, raising broad attention recently. 
At first, the SOTA approaches \cite{9696362,9636680} develop tracking-oriented low-light enhancers for nighttime UAV tracking, using leading-edge Siamese trackers \cite{8954116,9991169,9577739}. However, this kind of approach has a limited connection with tracking tasks, and straightforward insertion tracking models hardly learn to reduce the distribution gap at the feature level.
Due to the ability to reduce domain disparity and transfer knowledge from the source domain to the target domain, domain adaptation has been employed for various vision tasks~\cite{9578016,9577711}. UDAT \cite{9879981} brings unsupervised domain adaptation in nighttime UAV tracking for the first time, improving the tracker performance. Nevertheless, the one-step adaptation paradigm performs unstably when facing common LR object challenges in nighttime UAV tracking. The negative illumination conditions seriously weaken the detail information of the LR object, blurring its edge contour. Additionally, these adverse light conditions exacerbate nighttime noise interference. It is hard for the one-step adaptation paradigm to perceive and extract low-resolution object features directly.

\begin{figure*}[t]
  \centering
   \includegraphics[width=1.0\linewidth]{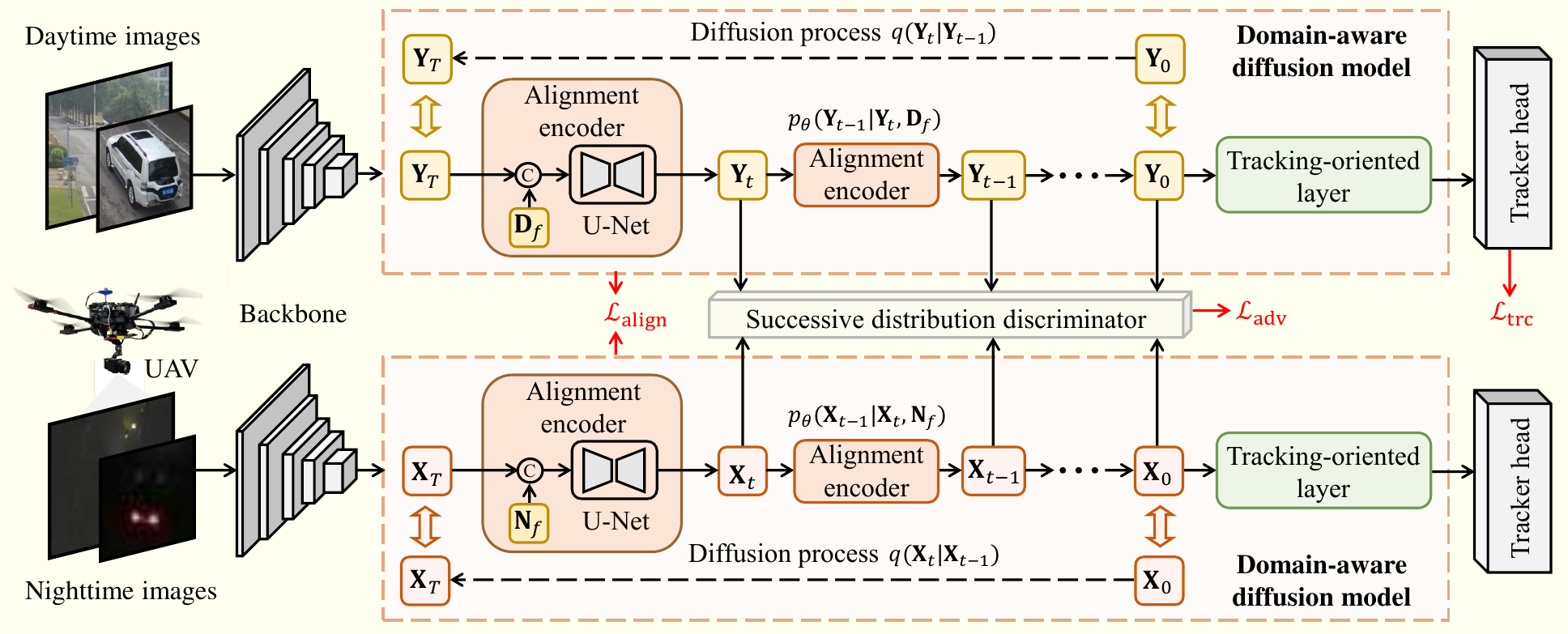}
   	\setlength{\abovecaptionskip}{-10pt}
   \caption{Overview of the proposed DaDiff. \textit{Domain-aware diffusion model} with \textit{alignment encoder} is employed to narrow feature distribution discrepancy successively, achieving the feature alignment for nighttime UAV tracking. \textit{Tracking-oriented layer} is developed to closely connect with the tracking tasks. \textit{Successive distribution discriminator} is trained to distinguish features between the daytime and the nighttime gradually. Best viewed in color. }
   \centering
   \label{2}
   	\vspace{-15pt}
\end{figure*}

\subsection{Diffusion models}

As a pioneering work, DDPM \cite{ho2020denoising} 
represents a unique class of variable models that leverage a Markov chain to transition from a noise distribution to a data distribution. Based on it, DDIM \cite{song2020denoising} adopts smaller sampling steps to speed up the generation process, with the characteristic of generating deterministic samples from random noise.
Recently, diffusion probabilistic models
have achieved SOTA performance in reconstructing the object information for LR images \cite{gao2023implicit,li2022srdiff,yue2024resshift,9887996}. S. Gao \textit{et al.}~\cite{gao2023implicit}
propose an implicit diffusion model for high-fidelity continuous LR image information enhancement. H. Li \textit{et al.}~\cite{li2022srdiff} introduce the diffusion probabilistic model into image super-resolution, handling the over smoothness and model collapse. Furthermore, Z. Yue \textit{et al.}~\cite{yue2024resshift} reduce the number of diffusion steps and eliminate the need for post-acceleration during inference, thereby realizing efficient LR image information recovery with the diffusion model. C. Saharia \textit{et al.} \cite{9887996} utilize the denoising diffusion probabilistic models to strengthen the detail information of LR images via repeated refinement. Despite significant development, diffusion models for nighttime UAV tracking have not been researched. 
Moreover, because diffusion models are trained independently, seamless integration into downstream tracking tasks remains elusive. Therefore, an effective diffusion model-based alignment framework for nighttime UAV tracking is urgently required.

\section{Proposed method}
In this section, the detailed structure of this work is described, as shown in Fig. \ref{2}. Throughout the training process, the proposed DaDiff aligns the features produced by the tracker backbone. In this procedure, adversarial learning successively reduces the gap between daytime and nighttime feature distributions at each diffusion timestep. By using this simple but effective alignment method, trackers can attain equal levels of stability and accuracy for night situations as they can during the daytime.

\subsection{Feature extraction network}

Siamese network feature extraction typically consists of two branches, the search branch and the template branch. By using the same backbone network, both branches extract feature maps from the template patch \textbf{T} and the search patch \textbf{S}, namely  $\mathcal{F}$(\textbf{T}) and  $\mathcal{F}$(\textbf{S}), by adopting an identical backbone network. Typically, trackers use the features of the last block or blocks for classification and regression.

\noindent\textit{\textbf{Remark 1:}} Since both $\mathcal{F}$(\textbf{T}) and $\mathcal{F}$(\textbf{S}) of daytime and nighttime will pass through the weight-share DaDiff and the discriminator, the following introduction uses the nighttime features $\textbf{N}_f$ as an example for clarity.

\subsection{Domain-aware diffusion model}
\noindent\textbf{Alignment encoder.} Diffusion models \cite{song2020denoising} are probabilistic models designed to learn a data distribution $p_{\theta}(\textbf{X}_{t-1}|\textbf{X}_{t})$ by gradually denoising a normally distributed variable, which corresponds to learning the reverse process of a fixed Markov chain of length $T$. The features extracted by the feature extraction network are input into the diffusion models to generate the version of the corresponding daytime distribution. 
\begin{figure}[t]
  \centering
  
   \includegraphics[width=1.0\linewidth]{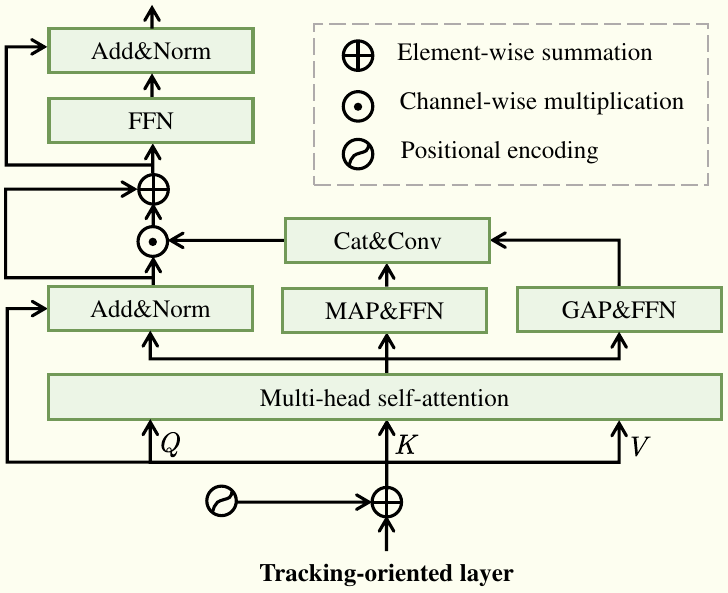}
  \vspace{-20pt}
   \caption{Detailed workflow of \textit{Tracking-oriented layer}. With the powerful information integration ability of Transformer \cite{vaswani2017attention} and the internal information exploration, \textit{Tracking-oriented layer} can integrate the effective domain-aware information of aligned LR object features, closely collaborating with the tracking tasks. }
   \centering
   \label{3}
   	\vspace{-20pt}
\end{figure}
Specifically, in the forward diffusion process, the noise is gradually added to the data  $\textbf{X}_t\sim{q(\textbf{X}_t|\textbf{X}_{t-1})}$ in $T$ steps with pre-defined value schedule $\alpha_t$:
\begin{equation}
\begin{aligned}
q(\textbf{X}_{t}|\textbf{X}_{t-1}) &= \mathcal{N}(\textbf{X}_{t};\sqrt{1-\beta_t}\textbf{X}_{t-1}, \beta_t\textbf{I})\quad ,
\end{aligned}
\end{equation}
where $\beta_t= 1-\alpha_t/\alpha_{t-1}$. 
A notable characteristic of diffusion models is that $\textbf{X}_{t}$ at an arbitrary time-step $t$ can be sampled from $\textbf{X}_{0}$ as:
\begin{equation}
\begin{aligned}
\textbf{X}_{t} &= \sqrt{\alpha_{t}}\ \textbf{X}_{0} + \sqrt{1-\alpha_t}\epsilon \quad ,
\end{aligned}
\end{equation}
where $\epsilon\sim{\mathcal{N}(0, \textbf{I})}$ is a noise variable.
While the reverse process is a process of noise removal. This process starts from random noise and gradually denoises to generate a real sample $p_{\theta}(\textbf{X}_{t-1}|\textbf{X}_t)$ according to the true distribution of each step of the reverse process. The proposed DaDiff utilizes the denoising process to achieve the day/night feature alignment, successively enhancing the diminished object information due to adverse illumination conditions. Thereby the reverse process is also the process of generating data:
\begin{equation}
\begin{aligned}
&p_{\theta}(\textbf{X}_{0:T}) = \mathcal{N}(\textbf{X}_{0};0, \textbf{I})\prod \limits_{t=1}^T p_{\theta}(\textbf{X}_{t-1}|\textbf{X}_{t}) \quad,\\[1mm]
&p_{\theta}(\textbf{X}_{t-1}|\textbf{X}_{t}) = \mathcal{N}(\textbf{X}_{t-1};\mu_{\theta}, \sigma_{\theta}^2\textbf{I}) \quad,
\end{aligned}
\end{equation}
where $\mu_{\theta}$ and $\sigma_{\theta}$ are parameters of the Gaussian distribution predicted by model $p_{\theta}$.
This process can be further interpreted as an equally weighted sequence of auto-encoders $\epsilon_\theta(\textbf{X}_t, t); t = 1, . . . ,T$, which are trained to predict a denoised variant of their input $\textbf{X}_t$, where $\textbf{X}_t$ is a noisy version of $\textbf{X}_0$. The process to obtain $\textbf{X}_{t-1}$ can be expressed by:
\begin{equation}
\begin{aligned}
&\hat{\textbf{X}}_{0} = \frac{\textbf{X}_{t}-\sqrt{1-\alpha_t}\epsilon_\theta(\textbf{X}_{t},t))}{\sqrt{\alpha_t}}\quad , \\[1mm]
&\textbf{X}_{t-1} = \sqrt{\alpha_{t-1}}\ \hat{\textbf{X}}_{0}+\sqrt{1-\alpha_{t-1}}\epsilon_\theta(\textbf{X}_{t},t) \quad ,
\end{aligned}
\end{equation}
then we can gradually get the desired data distribution by eliminating the noise predicted in each step. The corresponding objective $\mathcal{L}_{\textrm{dm}}$ can be simplified to:
\begin{equation}
\mathcal{L}_{\textrm{dm}} = \mathbb{E}_{\textbf{X}_0, \epsilon\sim\mathcal{N}, t}[\left\Vert\epsilon-\epsilon_\theta(\textbf{X}_t, t)\right\Vert^2_2]
\quad,
\end{equation}
where $\epsilon\sim{\mathcal{N}(0, \textbf{I})}$ is a noise variable and $t$ is uniformly sampled from \{1, . . ., $T$\}.

To meet the generation needs of specific tasks, the conditional mechanism is introduced into diffusion models~\cite{9878449}. Similar to other types of generative models~\cite{9555209}, diffusion models are in principle capable of modeling conditional distributions of the form $p_\theta(\textbf{X}_{t-1}|\textbf{X}_{t}, \textbf{N}_f)$. This can be implemented with a conditional alignment encoder $\epsilon_\theta(\textbf{X}_t, \textbf{N}_f, t)$ and paves the way to controlling the synthesis process through inputs $\textbf{N}_f$. Therefore, the proposed DaDiff concatenates $\textbf{X}_t$ with the flexible latent condition $\textbf{N}_f$ to augment the generation capabilities of tracking-specific distribution. Generally, $\epsilon_\theta(\textbf{X}_t, \textbf{N}_f, t)$ and $T$ timesteps are trained by a simplified objective $\mathcal{L}_{\textrm{align}}$:
\begin{equation}
\mathcal{L}_{\textrm{align}} = \mathbb{E}_{\textbf{X}_0, \textbf{N}_f, \epsilon\sim\mathcal{N}, t}[\left\Vert\epsilon-\epsilon_\theta(\textbf{X}_t, \textbf{N}_f, t)\right\Vert^2_2]\quad,
\end{equation}
where $\textbf{X}_t$ is a linear combination of data $\textbf{X}_0$ and noise $\epsilon$ by diffusion process. While during inference, the corresponding reverse generative Markov chain produces the expected data distribution $p_\theta(\textbf{X}_{t-1}|\textbf{X}_{t}, \textbf{N}_f)$ by denoising process. Afterward, the denoised image features are input into the tracking-oriented layer to integrate the object information, closely connecting with the tracking tasks.

\noindent\textit{\textbf{Remark 2:}} Through the successive denoising of nighttime features $\textbf{N}_f$, aligned image features can be generated stably and controllably. Thereby it is able to handle the common LR object challenges in the night scenes effectively, especially in the interference of adverse illumination conditions. 

\noindent\textbf{Tracking-oriented layer.} Diffusion models are difficult to directly collaborate with the tracking task due to their fixed training paradigm. Therefore, this work develops a tracking-oriented layer to integrate the domain-aware information of LR objects, bridging the aligned feature generation and the tracking process. In consideration of the strong modeling capability of the Transformer \cite{vaswani2017attention} for long-range inter-independencies, the tracking-oriented layer applies a Transformer structure, as shown in Fig. \ref{3}. The aligned features $\textbf{N}_f^{a}$ are obtained after this layer. Specifically, the denoised results $\textbf{X}_0$ are reshaped to $\textbf{X}_0^{a}$ before encoding. Subsequently, the input of this layer $\textbf{X}_0^{b}$ can be obtained by supplementing with a learnable positional encoding. The subsequent process can be expressed by:
\begin{equation}
\begin{aligned}
\textbf{X}_0^{c} &= \mathrm{Norm}(\mathrm{mAtt}(\textbf{X}_0^{b}) + \textbf{X}_0^{b})\quad,\\[1mm]
\mathcal{W} &= \mathrm{Conv}(\mathrm{Cat}(\mathrm{GAP}(\textbf{X}_0^{c}), \mathrm{MAP}(\textbf{X}_0^{c}))\quad,\\[1mm]
\textbf{X}_0^{d} &= \textbf{X}_0^{c} + \gamma_1 \ast \mathcal{W} \ast \textbf{X}_0^{c}\quad,\\[1mm]
\textbf{N}_f^{a} &= \mathrm{Norm}(\mathrm{FFN}(\textbf{X}_0^{d})+\textbf{X}_0^{d})\quad,
\end{aligned}
\end{equation}
where $\mathrm{mAtt}$ shows the mult-head self-attention. $\textbf{X}_0^{c}$, $\textbf{X}_0^{d}$ are intermediate variables and $\mathcal{W}$ is a weight matrix. $\mathrm{GAP}$ and $\mathrm{MAP}$ represent the global average pooling and max average pooling, thoroughly investigating the latent spatial information. $\mathrm{Norm}$ indicates layer normalization. In addition, $\mathrm{FFN}$ denotes the fully connected feed-forward network, which comprises of two linear layers separated by a ReLU. Besides, $\gamma_1$ and $\ast$ represent a learning weight and the channel-wise multiplication respectively. The final output is reshaped to the original size. 

\noindent\textit{\textbf{Remark 3:}} By virtue of superior information integration of Transformer, the proposed tracking-oriented layer is adequate to integrate the effective domain-aware information of aligned LR object features, thereby closely connecting the diffusion models with the tracking tasks. 
\begin{figure}[t]
  \centering
   \includegraphics[width=1.0\linewidth]{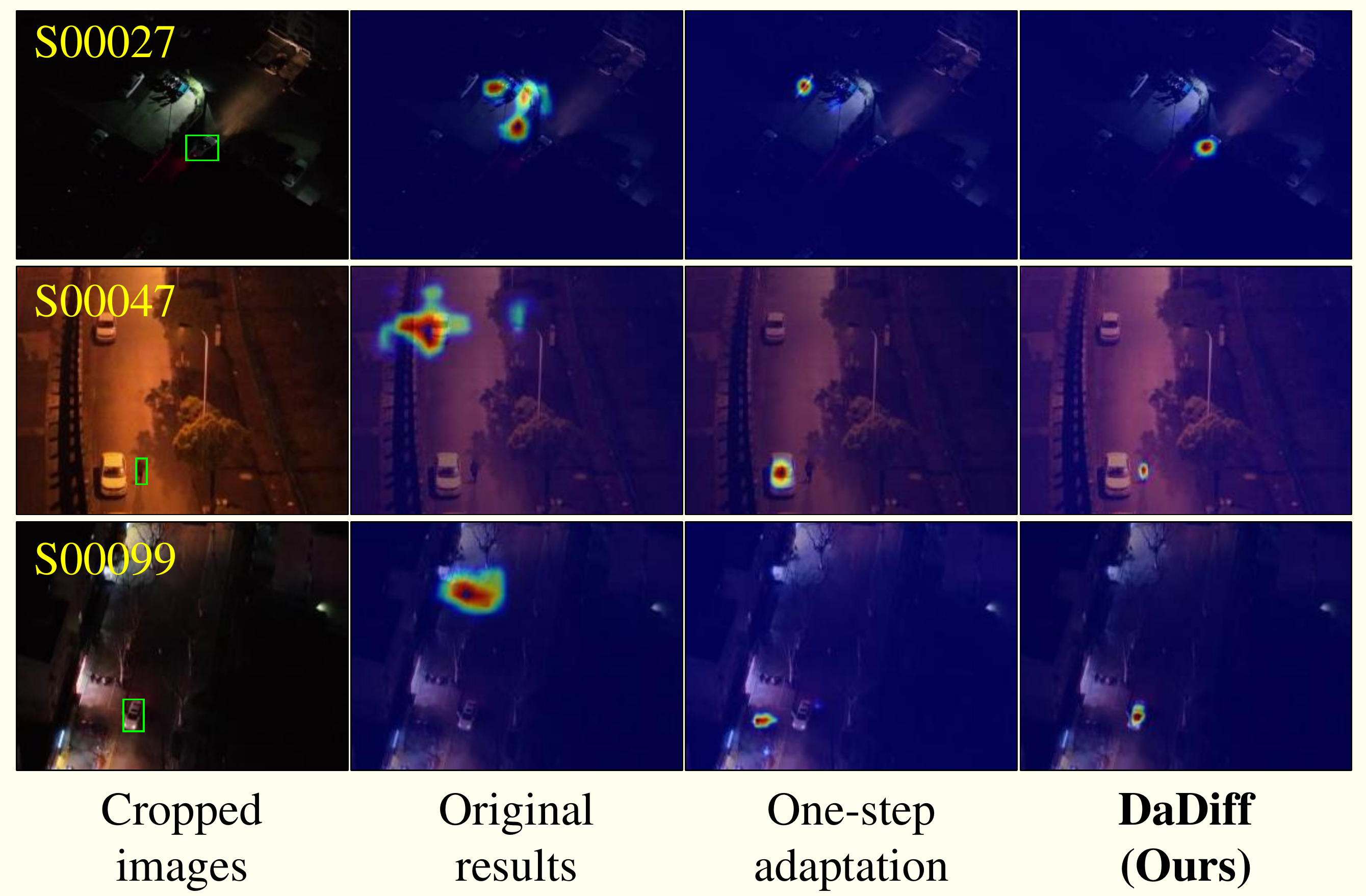}
\vspace{-20pt}
   \caption{ Visual comparison of confidence maps generated by the Baseline, the one-step adaptation paradigm, and the proposed DaDiff. Target objects are marked by {\color{green}green} boxes. The Baseline and the one-step adaptation paradigm struggle to extract robust LR object features in the interference of adverse illumination conditions. DaDiff stably and controllably aligns the image features by day/night domain awareness and applying the successive alignment strategy. }
   \centering
   \label{4}
   	\vspace{-15pt}
\end{figure}

\noindent\textbf{Successive distribution discriminator.}
The proposed DaDiff framework is trained in an adversarial learning manner. A successive distribution discriminator \cite{9879981} is applied to distinguish the different feature distributions at each diffusion timestep. Thereby DaDiff can step-by-step align the nighttime features with the daytime, thus handling the LR object challenges for nighttime UAV tracking, especially in the interference of adverse illumination conditions. In every diffusion process, the successive distribution discriminator $\mathrm{D}$ judges whether the features are from day or night. The adversarial optimization objective can be described as follows:
\begin{equation}
\begin{aligned}
\mathcal{L}_{\textrm{adv}} &=\sum_{t = 1}^{T}(\mathrm{D}(\textbf{X}_{t}) - l_d)^2 \quad ,
\end{aligned}
\end{equation}
where $t$ refers to the diffusion timestep. Besides, $l_d$ denotes the label for the daytime features, which has the same size as the output of $\mathrm{D}$. 

\begin{algorithm}[t]
\setlength{\partopsep}{2pt}
\setlength{\topsep}{2pt}
	\renewcommand{\algorithmicrequire}{\textbf{Input:}}
	\renewcommand{\algorithmicensure}{\textbf{Output:}}
	\caption{Domain-aware diffusion model}
	\label{alg1}
	\begin{algorithmic}[1]
        \REQUIRE nighttime features $\textbf{N}_f$
		\STATE  $\textbf{X}_t\sim{q(\textbf{X}_t|\textbf{X}_{t-1})}$
        \FOR{$t=T,...,1$}
        \STATE  $\hat{\textbf{X}}_{0} = \frac{\textbf{X}_{t}-\sqrt{1-\alpha_t}\epsilon_\theta(\textbf{X}_{t},\textbf{N}_f,t))}{\sqrt{\alpha_t}}$
        \STATE  $\textbf{X}_{t-1} = \sqrt{\alpha_{t-1}}\ \hat{\textbf{X}}_{0}+\sqrt{1-\alpha_{t-1}}\epsilon_\theta(\textbf{X}_{t},\textbf{N}_f,t)$
        \ENDFOR
		\STATE Send $\textbf{X}_{0}$ into tracking-oriented layer to generate $\textbf{N}_f^{a}$
		\ENSURE  aligned features $\textbf{N}_f^{a}$
	\end{algorithmic}  
\end{algorithm}

\noindent\textit{\textbf{Remark 4:}} DaDiff adopts the successive alignment strategy to generate the aligned features, more stable and controllable than the one-step adaptation paradigm. Thereby it can perceive and extract the robust LR object features in nighttime UAV scenes through gradual denoising. The superior performance of the proposed framework has been shown in Fig.~\ref{4}, using Grad-Cam~\cite{8237336}. Moreover, Algorithm \ref{alg1} displays the complete inference procedure of DaDiff.

\subsection{Tracker head}

Following the feature alignment, the tracker head predicts the tracked object's location using classification and regression. In the daytime training phase, the classification and regression loss $\mathcal{L}_{\textrm{trc}}$ are applied to connect DaDiff with the tracking task, assuring the trackers' normal tracking capacity. The applied tracking loss is commensurate with the baseline trackers without change. In conclusion, the total training loss for the proposed framework is defined as:
\begin{equation}
\begin{aligned}
\mathcal{L}_{\textrm{total}} &=\lambda_1\mathcal{L}_{\textrm{trc}} + \lambda_2\mathcal{L}_{\textrm{adv}} + \lambda_3\mathcal{L}_{\textrm{align}}\quad ,
   \end{aligned}
\end{equation}
where $\lambda_{1}$, $\lambda_{2}$, and $\lambda_{3}$ are the coefficients to balance the contributions of each loss, respectively.

\section{NUT-LR benchmark}

\begin{figure}[t]
  \centering
   \includegraphics[width=1.0\linewidth]{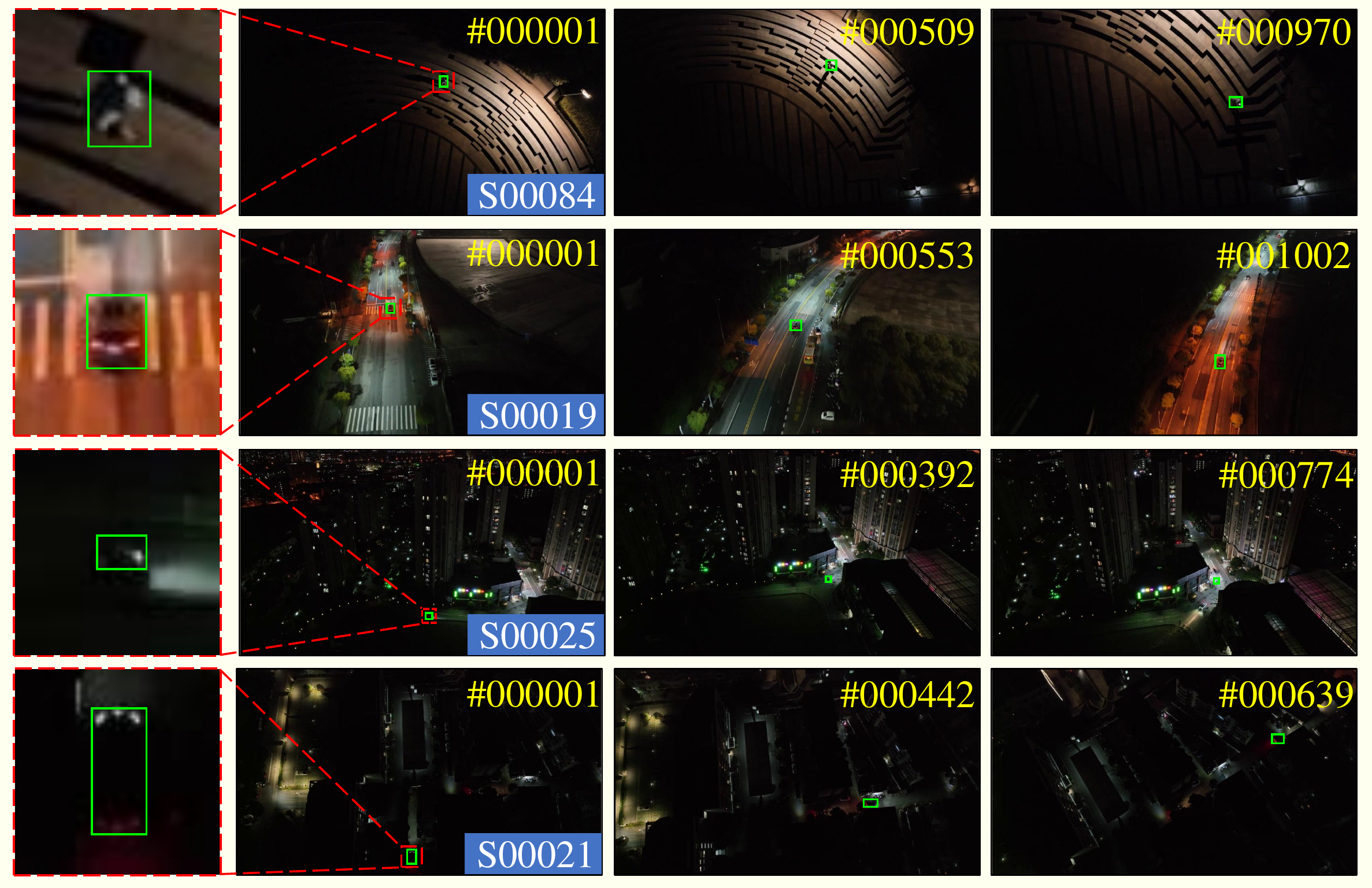}
   	\vspace{-20pt}
    
   \caption{Typically frames of selected sequences from NUT-LR. The {\color{green}green} boxes mark the tracked objects and the {\color{red}red} dotted boxes are the enlarged target areas for a clear view of the tracked LR objects. While the bottom-right corner of the image displays the sequence name and the top-right one shows the frame number. }
   \centering
   \label{5}
   	\vspace{-15pt}
\end{figure}

This work develops a nighttime UAV tracking dataset, namely NUT-LR, to evaluate the nighttime tracking performance comprehensively, especially for LR objects. Compared with the nighttime UAV tracking benchmarks \cite{yao2023sam,Li_2021_ICRA}, NUT-LR provides a dedicated dataset covering LR objects in various adverse light scenes highly related to the practical applications, as shown in Fig.~\ref{5}. 

\noindent\textit{\textbf{Remark 5:}} Referring to the authoritative public dataset~\cite{lin2014microsoft} and actual UAV tracking, the LR object is defined that the size of the target is less than 25×25 in NUT-LR.

\subsection{Data collection}

A classical UAV platform is applied to photograph images of NUT-LR in diverse evening views at 30 frames/s, such as highways, squares, bridges, and universities. The UAV tracks LR objects from an UAV perspective of more than 100 meters. Sequence categories include various objectives, \textit{e.g.}, cars, persons, groups, bikes, and motorcycles. Moreover, the proposed benchmark NUT-LR contains 100 nighttime UAV tracking sequences in total.

\subsection{Attributes}

The test sequences of NUT-LR are categorized into 10 various attributes to provide a thorough study of trackers, including aspect ratio change (ARC), background clutter (BC), camera motion (CM), fast motion (FM), occlusion (OCC), scale variation (SV), similar object (SOB), viewpoint change (VC), illumination variation (IV), and low ambient intensity (LAI) \cite{9879981}. Tracking LR objects under these attributes is more challenging than general objects. Due to the few pixels in the nighttime images, the detail information of the LR object is seriously weakened by adverse illumination conditions, such as IV and LAI. Therefore, NUT-LR can promote the tracker designs with additional challenges. 

\noindent\textit{\textbf{Remark 6:}} SOTA trackers are evaluated on the proposed benchmark and the results demonstrate that existing trackers and the one-step adaptation paradigm hardly provide sufficient performance when confronting LR objects in nighttime UAV tracking.

\begin{table*}[!b]
\renewcommand{\arraystretch}{0.85}
\footnotesize
  \centering
  \vspace{-10pt}
  \caption{Performance comparison between baseline trackers and DaDiff. $\Delta$ indicates percentage increases brought by DaDiff. Prec. and Succ. mean the precision and the success rate. DaDiff has improved nighttime UAV tracking performance significantly.}
  \vspace{-10pt}
    \label{tab_bsl}
  \setlength{\tabcolsep}{3.8mm}{
  \colorbox{table_c}{
    \begin{tabular}{cccccccccccccccccccc}
    \toprule
    Benchmark & \multicolumn{3}{c}{NUT-LR} & \multicolumn{3}{c}{NUT-L} & \multicolumn{3}{c}{UAVDark70} \\
    \midrule
    Metric & Succ. & Norm. Prec. & Prec. & Succ. & Norm. Prec. & Prec.& Succ. & Norm. Prec. & Prec.\\
    \midrule
    SiamGAT & 0.508 & 0.569 & 0.740 & 0.412 & 0.474 & 0.540 & 0.487 & 0.583 & 0.655 \\
    DaDiff-GAT & 0.531 & 0.585 & 0.765 & 0.449 & 0.508 & 0.584 & 0.511 & 0.610 & 0.698 \\
    $\Delta_{\mathrm{GAT}}$(\%) & \textbf{+4.5} & \textbf{+2.8} & \textbf{+3.4} & \textbf{+9.0} & \textbf{+7.2} & \textbf{+8.1} & \textbf{+4.9} & \textbf{+4.6} & \textbf{+6.6} \\
    \midrule
    SiamBAN & 0.507 & 0.540 & 0.753 & 0.350 & 0.398 & 0.466 & 0.489 & 0.570 & 0.677  \\
    DaDiff-BAN & 0.538 & 0.574 & 0.777 & 0.401 & 0.450 & 0.526 & 0.518 & 0.601 & 0.706\\
    $\Delta_{\mathrm{BAN}}$(\%) & \textbf{+6.1} &\textbf{+6.3} & \textbf{+3.2} & \textbf{+14.6} & \textbf{+13.1} & \textbf{+12.9} & \textbf{+5.9} & \textbf{+5.4} & \textbf{+4.3}\\
    
    \bottomrule
    \end{tabular}%
  }}
\end{table*}%

\begin{table*}[!b]
\renewcommand{\arraystretch}{0.8}
\footnotesize
  \centering
  \vspace{-5pt}
  \caption{Comparison of the one-step adaptation paradigm and DaDiff. Norm., Prec., Succ., DA, and $\Delta$ indicate the normalization, the precision, the success rate, the domain adaptation, and the percentage increase, respectively. The tracker with DaDiff achieves superior tracking performance in all nighttime UAV tracking benchmarks.}
  \vspace{-10pt}
    \setlength{\tabcolsep}{3.4mm}{
    \colorbox{table_c}{
    \begin{tabular}{ccccccccccccccccccc}
    \toprule
    Benchmark & \multicolumn{3}{c}{NUT-LR} & \multicolumn{3}{c}{NUT-L} & \multicolumn{3}{c}{UAVDark70} \\
    \midrule
    Metric & Succ. & Norm. Prec. & Prec. & Succ. & Norm. Prec. & Prec.& Succ. & Norm. Prec. & Prec.\\
    \midrule
    SOTA one-step DA \cite{9879981} & 0.517 & 0.562 & 0.764 & 0.377 & 0.434 & 0.498 & 0.510 & 0.597 & 0.702  \\
    DaDiff-BAN & \textbf{0.538} & \textbf{0.574} & \textbf{0.777} & \textbf{0.401} & \textbf{0.450} & \textbf{0.526} & \textbf{0.518} & \textbf{0.601} & \textbf{0.706}\\
    \midrule
    $\Delta$(\%) & \textbf{+4.1} & \textbf{+2.1} & \textbf{+1.7} & \textbf{+6.4} & \textbf{+3.7} & \textbf{+5.6} & \textbf{+1.6} & \textbf{+0.7} & \textbf{+0.6} \\
    
    \bottomrule
    \end{tabular}%
  }}
  \label{tab_udat}%
\end{table*}%

\begin{figure*}[!ht]	
	\centering
	{
\colorbox{table_c}{
	\includegraphics[width=0.32\linewidth]{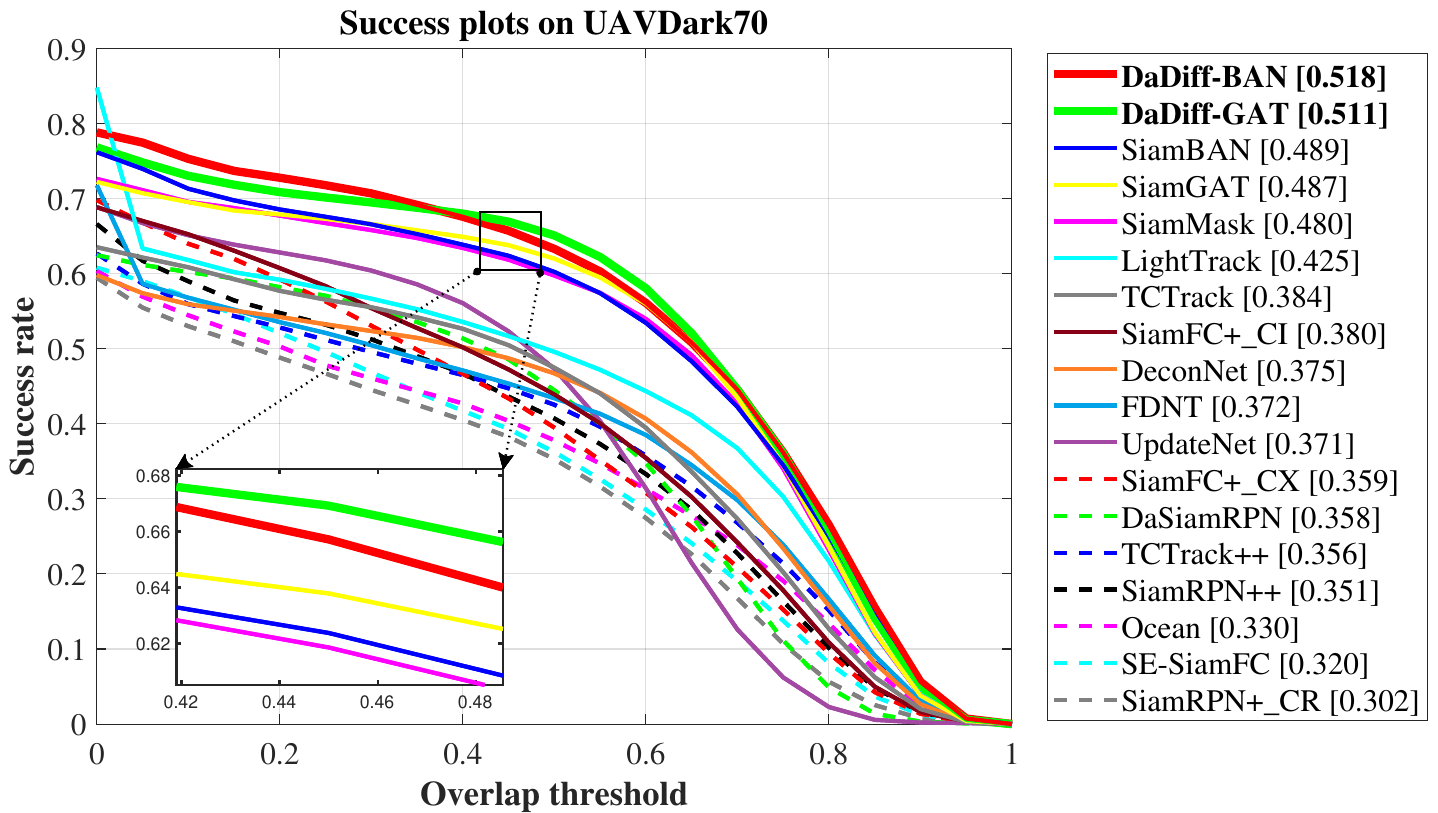}
	\includegraphics[width=0.32\linewidth]{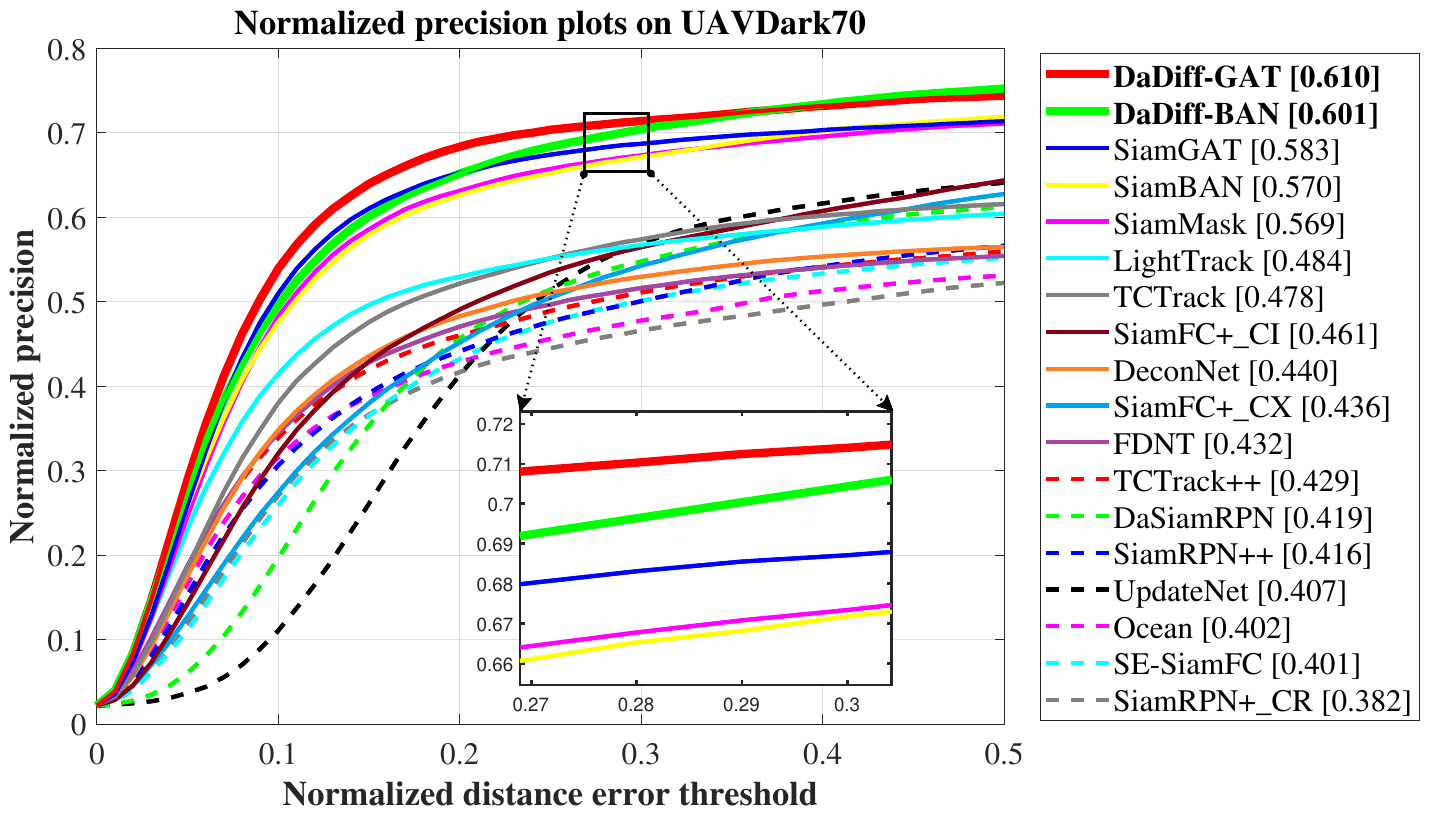}
   \includegraphics[width=0.32\linewidth]{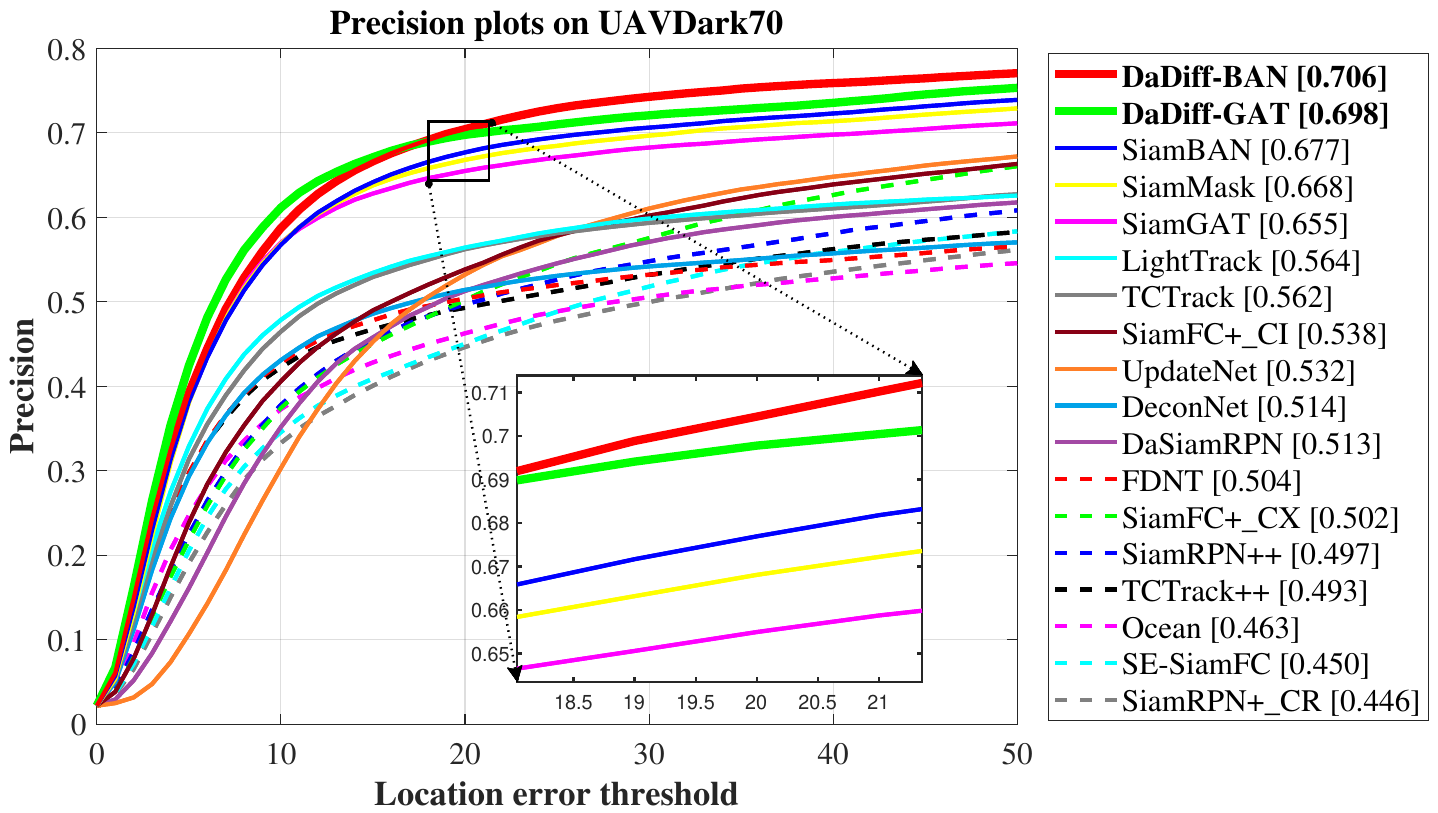}
}}
\\[0.01em]
{
\colorbox{table_c}{
	\includegraphics[width=0.32\linewidth]{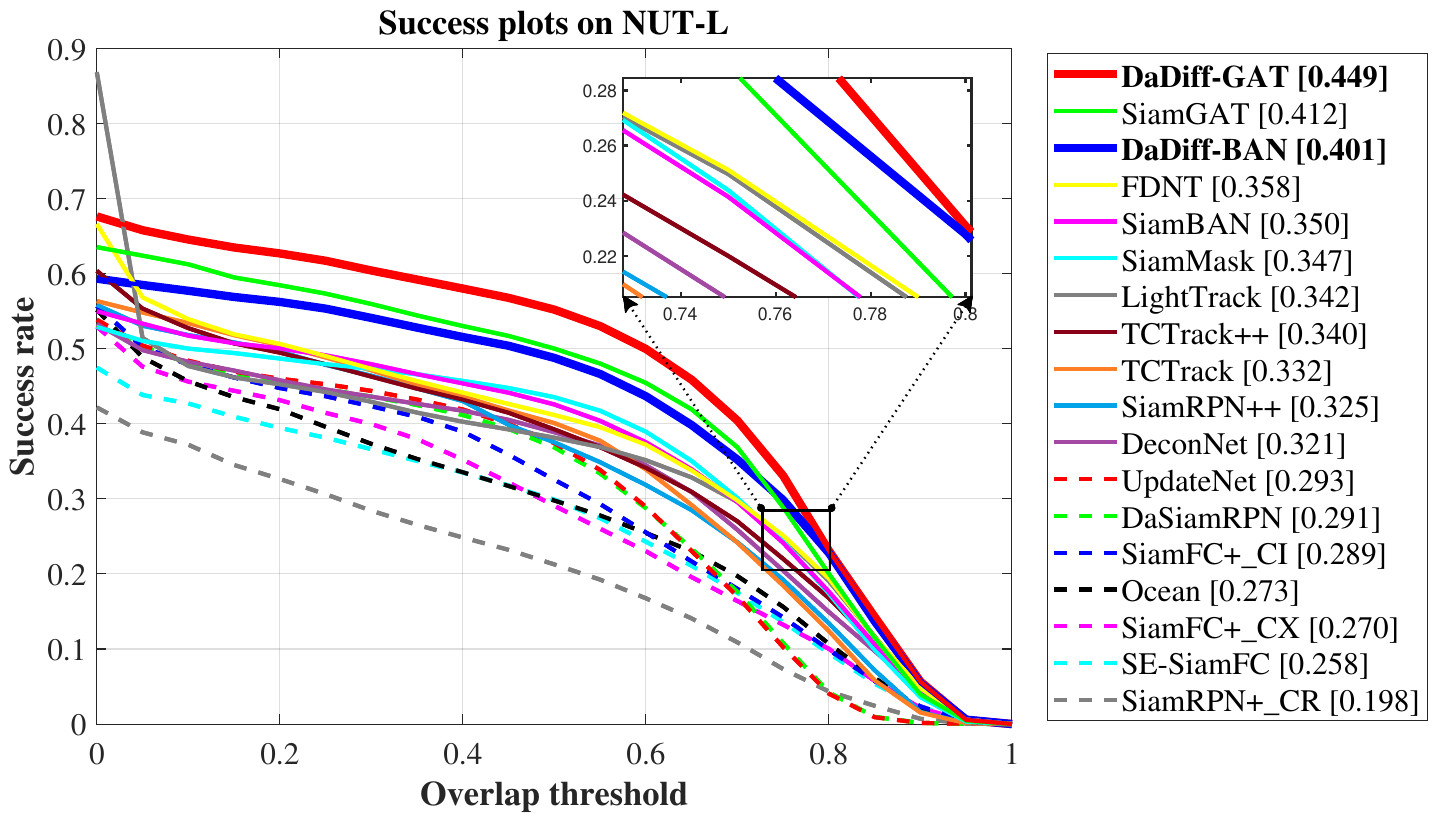}
	\includegraphics[width=0.32\linewidth]{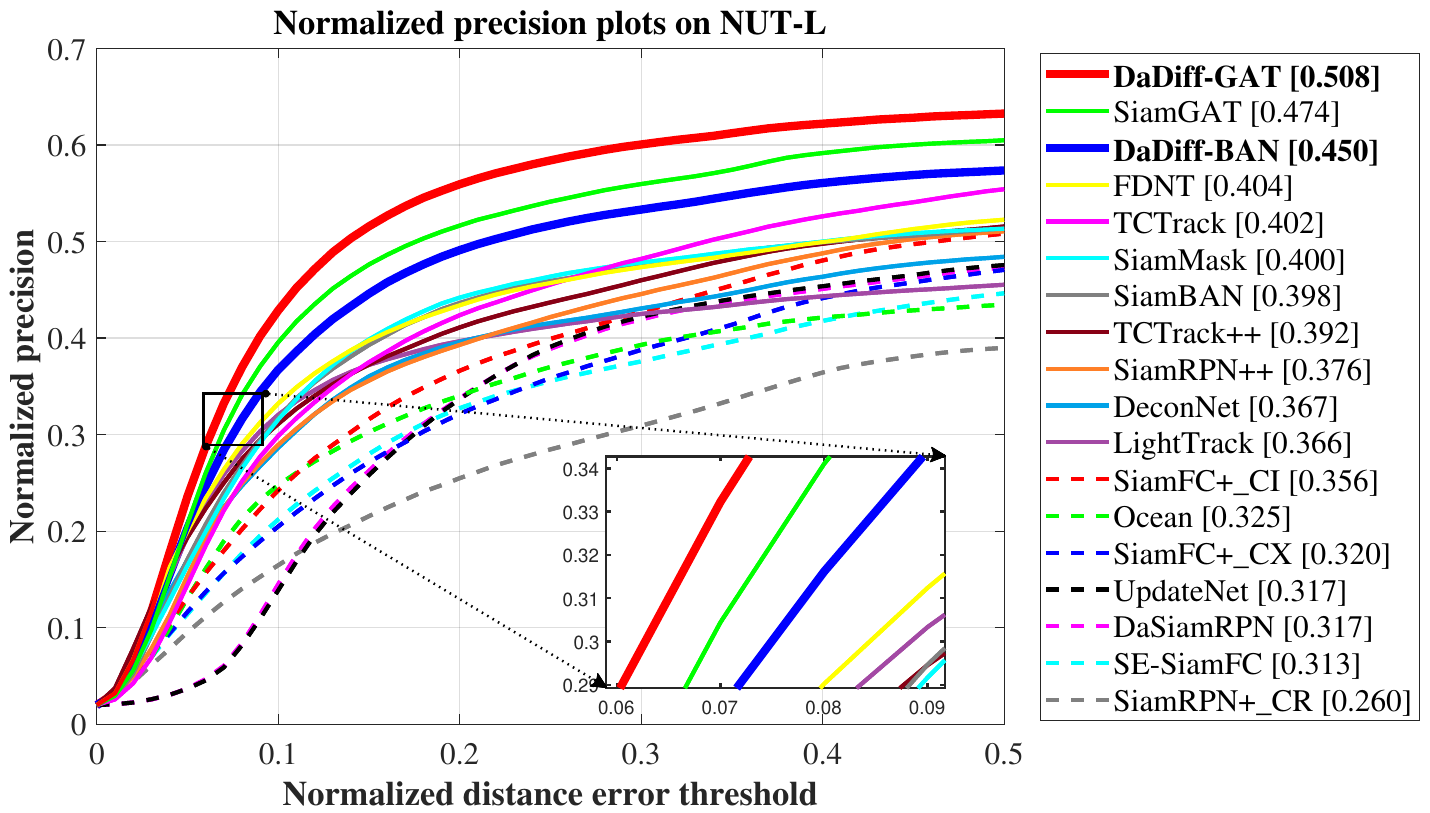}
   \includegraphics[width=0.32\linewidth]{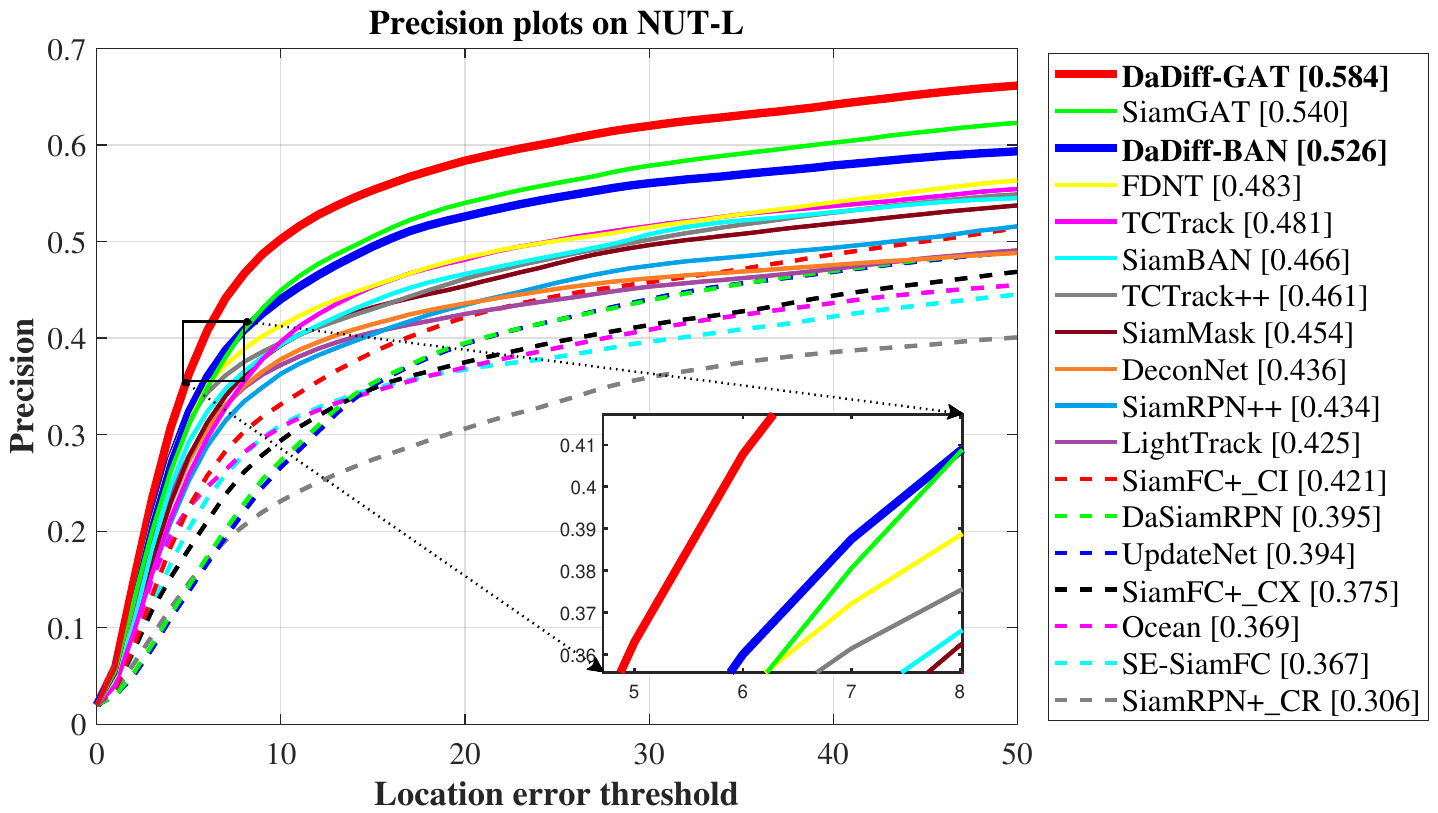}
}}\\[0.01em]
{
 \colorbox{table_c}{
	\includegraphics[width=0.32\linewidth]{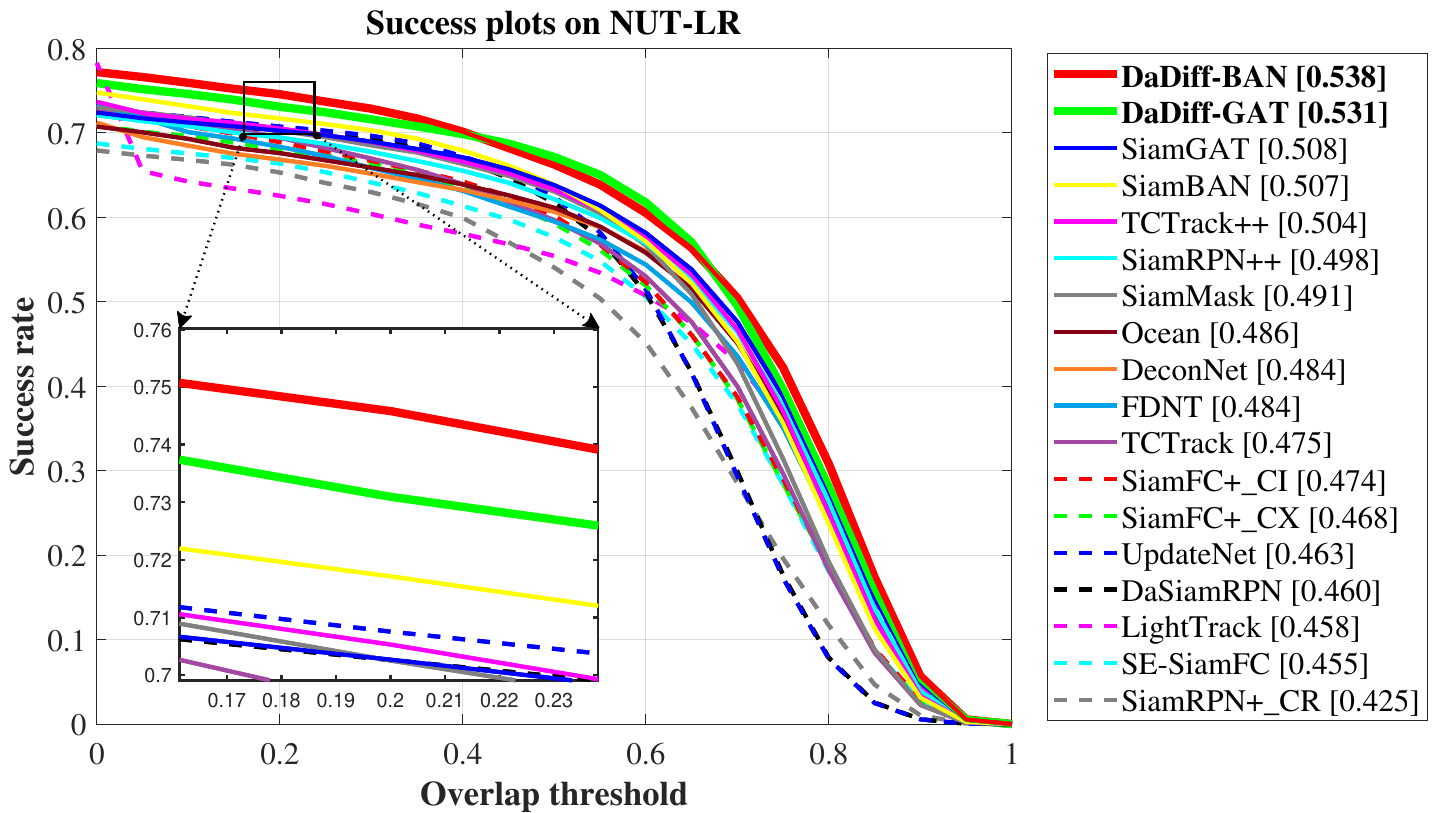}
	\includegraphics[width=0.32\linewidth]{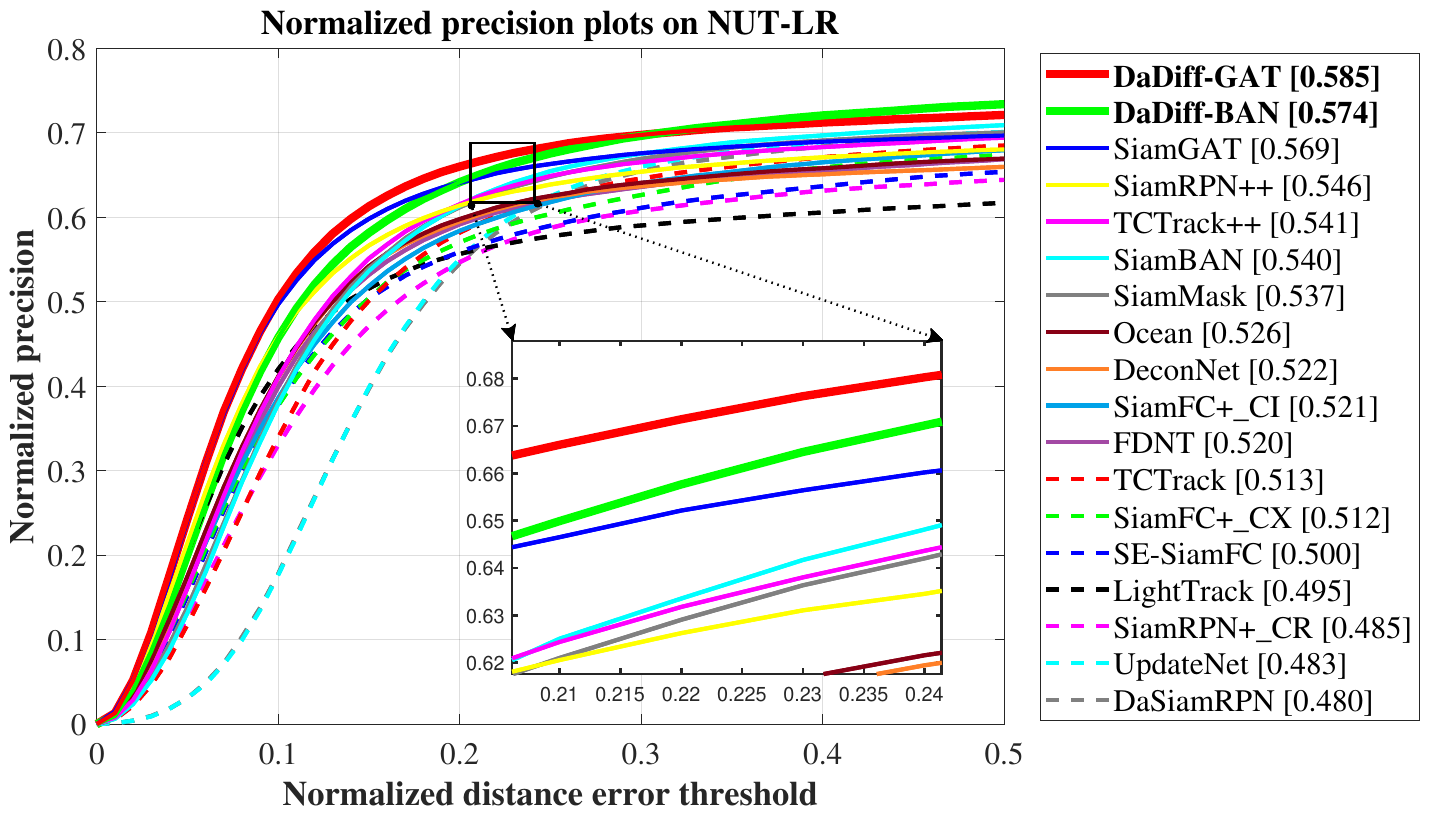}
   \includegraphics[width=0.32\linewidth]{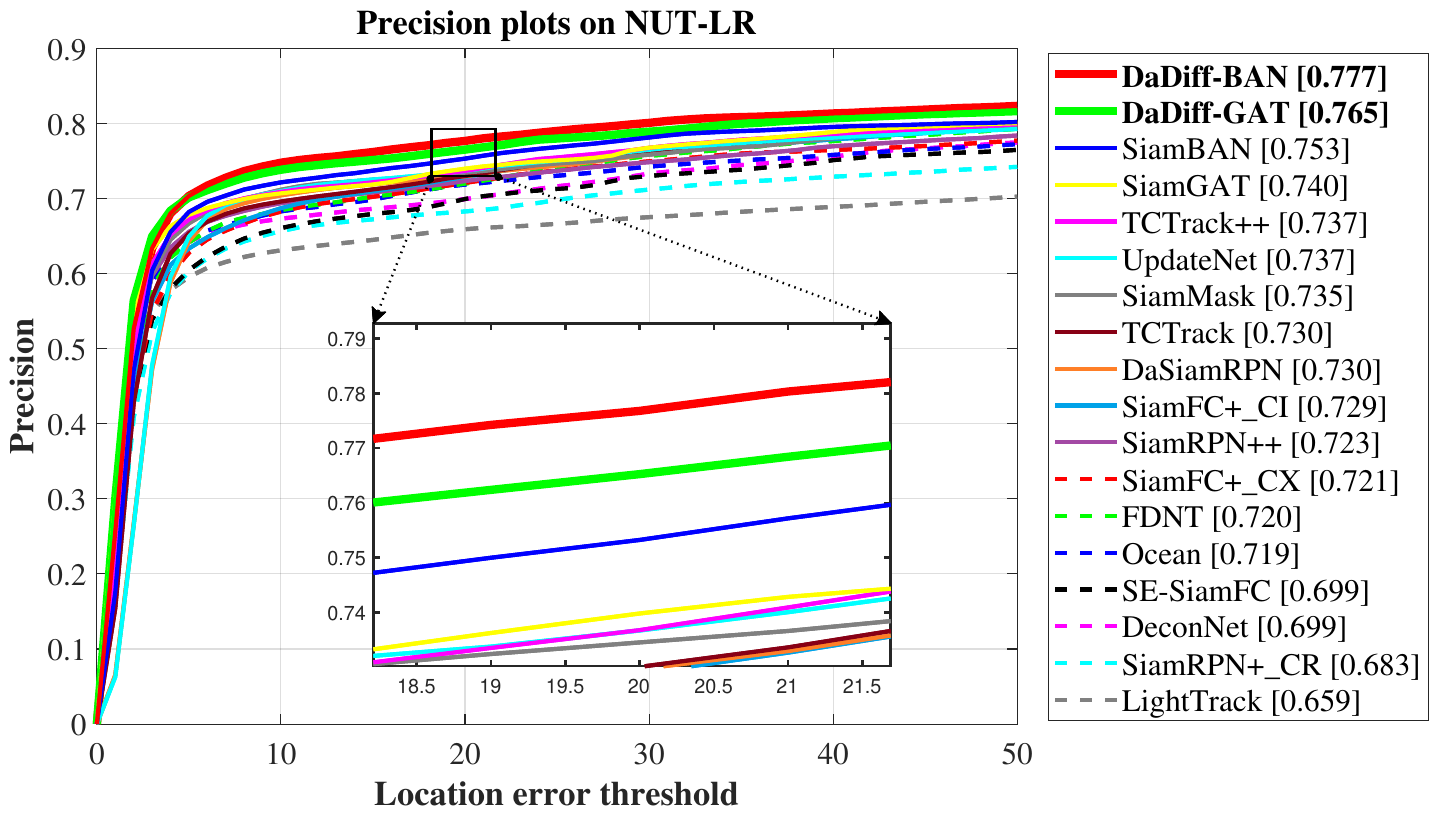}
}}
\vspace{-10pt}
	\caption
	{
		Overall performance of SOTA trackers and DaDiff on  UAVDark70 \cite{Li_2021_ICRA}, NUT-L \cite{yao2023sam}, and NUT-LR benchmarks. The evaluation results indicate
that the proposed method improves the tracking performance on all benchmarks.
	}
	\label{fig_per}
	\vspace{-15pt}
\end{figure*}

\section{Experiments}
\subsection{Implementation details}

The proposed DaDiff framework is implemented on an NVIDIA A100 Tensor Core GPU with PyTorch. The base learning rate of the successive distribution discriminator is set at 0.005 and decays according to the poly learning rate policy with a 0.8 power. While DaDiff adopts a base learning rate of 0.0015 and is optimized with the baseline tracker. There are 50 epochs throughout the whole training procedure. The SOTA trackers~\cite{9577739,9157457} are adopted as Baselines. Pre-trained tracking models on generic datasets \cite{8922619,lin2014microsoft,russakovsky2015imagenet} are used as the baseline models to accelerate convergence. For the sake of fairness, in the daytime training branch, only the tracking datasets~\cite{8922619,russakovsky2015imagenet} on which the pre-trained models were trained are used, and no additional daytime datasets are added. 
While in the nighttime training branch, the unlabeled benchmark NAT2021-train \cite{9879981} is applied for alignment training. 

\subsection{Evaluation metrics}
In one-pass evaluation metrics, precision, normalized precision, and success rate are key factors for evaluating tracker performance \cite{muller2018trackingnet}. The success rate is calculated by considering the intersection over union (IoU) between the actual and predicted bounding boxes. The success plot represents the fraction of frames where the IoU exceeds a preset threshold. Precision, on the other hand, is determined by measuring the center location error (CLE) between the predicted and actual locations. The precision plot visualizes the share of frames where the CLE falls within a specific range. Furthermore, normalized precision is obtained by normalizing precision across different sizes of the ground truth bounding box, aiming to eliminate the impact of varying object sizes on precision. The normalized precision plot is evaluated by calculating the area under the curve.

\subsection{Evaluation results}

16 SOTA trackers \cite{9981882,cao2023towards,9157457,8954116,9991169,9577739,bertinetto2016fully,yan2021lighttrack,9423458,hu2023siammask,9008117,8953458,zhang2020ocean} are tested on NUT-LR, together with the proposed DaDiff, to provide an extensive evaluation of trackers for nighttime UAV tracking and aid further research. For clarity, two trackers further trained by DaDiff are named DaDiff-GAT and DaDiff-BAN, respectively. Moreover, two challenging and authoritative public datasets, \textit{i.e.}, NUT-L \cite{yao2023sam} and UAVDark70 \cite{Li_2021_ICRA} are also served as the evaluation benchmarks. 

\noindent\textit{\textbf{Remark 7:}} For the justice, every compared tracker adopts the tracking model from the official code and all evaluation experiments are completed on the same platform.

\begin{table}[b]
\vspace{0pt}
	\centering
 \vspace{-15pt}
 \caption{Attribute-based evaluation of top 6 trackers on NUT-LR. The best two performances are respectively highlighted in the {\color{red}red} and {\color{green}green} colors. The trackers with DaDiff have improved the tracking performance of original trackers in different attributes.}
    \label{tab_att}
	\resizebox{\linewidth}{!}{
  \colorbox{table_c}{
	\begin{tabular}{lccccccccccccccccccccccccccccccccccccccccccccccccccccccccccccccccccc|c}
		 \toprule
			Attributes & \multicolumn{2}{c}{ARC} & \multicolumn{2}{c}{SV}  & \multicolumn{2}{c}{IV}\\
          \midrule
			Trackers & Succ. & Norm. Prec. & Succ. & Norm. Prec. & Succ. & Norm. Prec. \\
   \midrule
           TCTrack++ & 0.451 & 0.472 & 0.538 & 0.582 & 0.502 & 0.542   \\
           SiamRPN++ & 0.478 & 0.512 & 0.543 & 0.598 & 0.499 & 0.550  \\
           SiamGAT & 0.480 & 0.530 & 0.556 & \textbf{{\color{green}0.632}} & 0.516 & 0.579 \\
           SiamBAN & 0.483 & 0.512 & 0.547 & 0.585 & 0.508 & 0.541 \\
    \midrule
           \textbf{DaDiff-GAT} & \textbf{{\color{green}0.504}} & \textbf{{\color{red}0.545}} & \textbf{{\color{red}0.586}} & \textbf{{\color{red}0.653}} & \textbf{{\color{green}0.537}} & \textbf{{\color{red}0.594}} \\
           \textbf{DaDiff-BAN} & \textbf{{\color{red}0.506}} & \textbf{{\color{green}0.538}}& \textbf{{\color{green}0.581}} & 0.623 & \textbf{{\color{red}0.542}} & \textbf{{\color{green}0.581}} \\
		\bottomrule
	     \end{tabular}
	     }}
      \vspace{0pt}
\end{table}
\begin{table}[b]
\renewcommand{\arraystretch}{0.8}
    \small
	\centering
 \vspace{-15pt}
	\caption{Ablation study of various parts of the proposed framework on NUT-LR. $\Delta$ symbolizes the improvement
    over Baseline. Prec. and Succ. represent the precision and the success rate respectively.}
    \label{tab_abl}
	\resizebox{\linewidth}{!}{
  \colorbox{table_c}{
		\begin{tabular}{lcccccccccccccccccccc|c}
		 \toprule
		AE & TL & SD & Prec. & $\Delta_{prec}$(\%) & Succ. & $\Delta_{succ}$(\%) \\
		\midrule 
		-  & -  & - & 0.640  & - & 0.426  & - \\ 
            \checkmark  & -  & - & 0.638  & -0.3 & 0.431  & +1.2\\
            \checkmark  &  - & \checkmark & 0.654  & +2.2 & 0.439  & +3.1\\
            \checkmark  &  \checkmark & - & 0.658  & +2.8 & 0.442  & +3.8\\
            \midrule
            \checkmark  &  \checkmark & \checkmark & \textbf{0.677}  & \textbf{+5.8} & \textbf{0.452}  & \textbf{+6.1}\\
		\bottomrule
	     \end{tabular}
	    } }
\end{table}

\subsubsection{Overall performance}

\noindent\textbf{UAVDark70.} As illustrated in the top row of Fig. \ref{fig_per}, DaDiff trackers raise the performance of SiamBAN (0.489) and SiamGAT (0.487) by 5.9\% and 4.9\%.
It can be demonstrated that the proposed method has improved the tracking performance of the baseline trackers against different nighttime tracking challenges.

\noindent\textbf{NUT-L.} Results in the second row of Fig. \ref{fig_per} show the proposed DaDiff-BAN and DaDiff-GAT consistently achieve satisfactory results. Apart from LR objects, the proposed method can improve the tracking performance in various long-term nighttime scenes significantly. In success rate, DaDiff improves the baseline trackers by over 9\%.

\noindent\textbf{NUT-LR.} As indicated in the third row of Fig. \ref{fig_per}, DaDiff-BAN and DaDiff-GAT rank first two places with a large margin compared to their Baselines. A performance comparison of DaDiff and baseline trackers is reported in TABLE~\ref{tab_bsl}. In success rate, DaDiff-BAN (0.538) and DaDiff-GAT (0.531) raise the original SiamBAN (0.507) and SiamGAT (0.508) by 6.1\% and 4.5\%, respectively.

\noindent\textit{\textbf{Remark 8:}} The improvement brought by DaDiff attests to the efficacy of the proposed diffusion models-based alignment framework, particularly for tracking LR objects.

\subsubsection{Comparison with one-step adaptation paradigm}

To prove the alignment effect and robustness of the proposed DaDiff compared with the one-step adaptation paradigm, the previous SOTA method \cite{9879981} is used for evaluation. SiamBAN \cite{9157457} is selected as the baseline tracker. As shown in TABLE~\ref{tab_udat}, DaDiff is superior in all benchmarks. The superior results prove that DaDiff is competent for feature alignment in nighttime UAV tracking, especially for tracking LR objects.

\noindent\textit{\textbf{Remark 9:}}
To be fair, the compared previous SOTA approach employs the official code's pre-trained model.

\begin{figure}[t]
  \centering
   \includegraphics[width=0.98\linewidth]{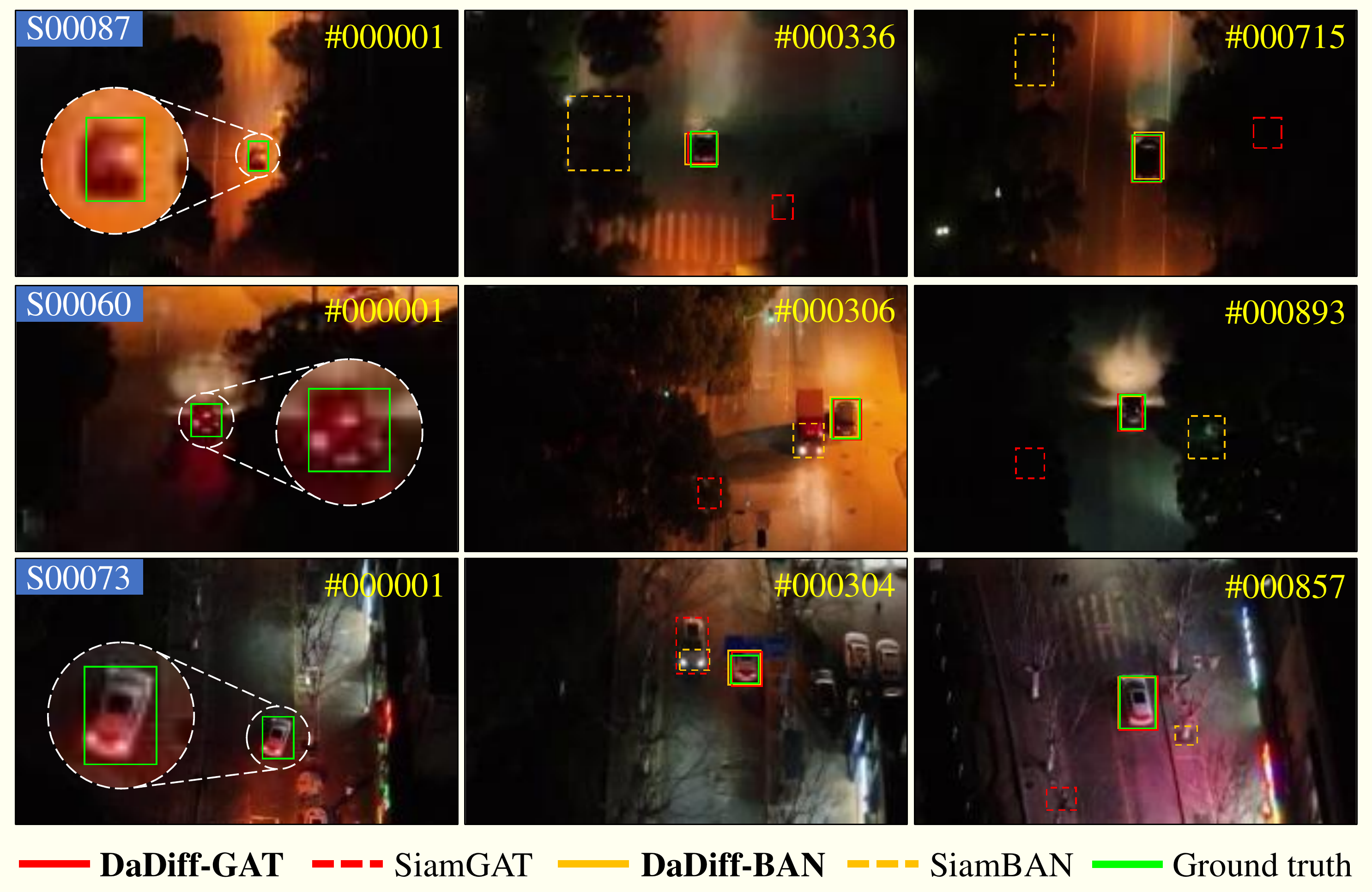}
   \vspace{-5pt}
    
   \caption{Qualitative comparison of SOTA trackers with and without DaDiff on NUT-LR. The cropped original image frames are blurred due to LR objects. When the baseline trackers lose nighttime objects, DaDiff improves the perception ability of trackers to successfully track these LR objects.}
   \centering
   \label{fig_qa}
   	\vspace{-15pt}
\end{figure}

\subsubsection{Attribute-based evaluation}

To exhaustively evaluate DaDiff when facing LR objects with various challenges, attribute-based comparisons are conducted on NUT-LR, as shown in TABLE \ref{tab_att}. The trackers with DaDiff achieve superior performance in comparison with other top 4 trackers. Specifically, DaDiff significantly improves the performance of tracking LR objects in attributes of ARC, SV, and IV. The satisfactory results demonstrate that DaDiff can gradually upgrade the detail information of LR objects in adverse illumination conditions.

\subsubsection{Qualitative evaluation}

Some typical nighttime tracking scenes and the performance of SOTA trackers in NUT-LR are displayed in Fig. \ref{fig_qa}. The baseline trackers fail to focus on LR objects in poor lighting, whereas DaDiff significantly improves the baseline trackers' capacity for nighttime perception, producing satisfactory nighttime tracking performance.

\subsection{Ablation study}

To verify the effectiveness of the proposed framework, comprehensive ablation studies are presented in this subsection. SiamBAN \cite{9157457} and NUT-LR are chosen as the Baseline and the evaluation benchmark, respectively. For clarity, we first introduce various modules, including alignment encoder (AE), tracking-oriented layer (TL), and successive distribution discriminator (SD). The results on TABLE \ref{tab_abl} show that AE slightly promotes nighttime tracking, with a slight upgrade in success rate but degradation in precision. It indicates that a lack of TL for bridging the feature alignment with the tracking tasks can lead to task mismatch. Therefore, AE can hardly learn the data distribution suitable for nighttime UAV tracking. When employing TL, performance on the nighttime tracking scenes obtains a 3.8\% boost in success rate, which verifies the effectiveness of the proposed tracking-oriented layer. Furthermore, SD increases the promotion brought by AE due to the successive feature distribution discrimination process. 
The results verify that the proposed various modules fairly assist the tracker in generating discriminative features from nighttime images, especially for LR object challenges.

\section{Conclusion}

This work proposes a novel progressive alignment paradigm, named domain-aware diffusion model (DaDiff), for nighttime UAV tracking, especially handling LR object challenges. Specifically,  an alignment encoder is developed to enhance the detail information of LR objects weakened by adverse illumination conditions. A tracking-oriented layer is developed to achieve close collaboration with the tracking tasks. To ensure the stability of the alignment effect, a successive distribution discriminator is applied for distinguishing the different feature distributions at each diffusion timestep. 
Detailed evaluation on nighttime tracking benchmarks shows the effectiveness and feature alignment ability of DaDiff. To summarize, we are confident that this work can contribute to the advancement of visual tracking at night and in other challenging environments, especially for LR objects.

\vspace{-5pt}
\section*{Acknowledgement}
This work is supported by the National Natural Science
Foundation of China (No. 62173249), the Natural Science
Foundation of Shanghai (No. 20ZR1460100), and the Innovation and Technology Commission of the HKSAR Government under the InnoHK initiative.










\bibliographystyle{IEEEtran}
\bibliography{root}

\end{document}